\documentclass{article}

\usepackage[final]{corl_2020} 

\usepackage{amsmath}
\usepackage{amsfonts}
\usepackage[inline]{enumitem}
\usepackage{graphicx}
\usepackage{multirow}
\usepackage{booktabs} 

\newlength\paramargin
\newlength\figmargin
\newlength\secmargin

\newlength\figwidth

\newcommand{\secref}[1]{Sec.~\ref{#1}}

\long\def\ignorethis#1{}

\newcommand{\myparagraph}[1]{\vspace{0.1in}\noindent\textbf{#1}}

\newcommand{\hreff}[2]{\href[pdfnewwindow=true]{#1}{\nolinkurl{#2}}}

\title{Tactile Object Pose Estimation from the First Touch with Geometric Contact Rendering}

\author{
  Maria Bauza, Eric Valls, Bryan Lim, Theo Sechopoulos, Alberto Rodriguez\\ 
  Massachusetts Institute of Technology \\
  \texttt{\{bauza,albertor\}@mit.edu} \\
}

\begin{document}
\maketitle

\begin{abstract}
In this paper, we present an approach to tactile pose estimation from the first touch for known objects.  
First, we create an object-agnostic map from real tactile observations to contact shapes. 
Next, for a new object with known geometry, we learn a tailored perception model completely in simulation.
%
%
To do so, we simulate the contact shapes that a dense set of object poses would produce on the sensor. 
Then, given a new contact shape obtained from the sensor output, we match it against the pre-computed set using the object-specific embedding learned purely in simulation using contrastive learning. 

This results in a perception model that can localize objects from a single tactile observation. 
It also allows reasoning over pose distributions and including additional pose constraints coming from other perception systems or multiple contacts.

We provide quantitative results for four objects. 
Our approach provides high accuracy pose estimations from distinctive tactile observations while regressing pose distributions to account for those contact shapes that could result from different object poses. 
We further extend and test our approach in multi-contact scenarios where several tactile sensors are simultaneously in contact with the object.

Website: \hreff{http://mcube.mit.edu/research/tactile_loc_first_touch.html}{mcube.mit.edu/research/tactile\_loc\_first\_touch.html}
\end{abstract}



\section{Introduction}
\vspace{-2mm}

\label{sec:introduction}

Robotics history sends a clear lesson: accurate and reliable perception is an enabler of progress in robotics. 
From depth cameras to convolutional neural networks, we have seen how advances in perception foster the development of new techniques and applications.
For instance, the invention of high-resolution LIDAR fueled self-driving cars,  
and the generalization capacity of deep neural networks has dominated progress in perception and grasp planning in warehouse automation~\citep{zeng_2017,milan2018semantic,schwarz2018fast}.
The long term goal of our research is to understand the key role that tactile sensing plays in that progress. 
In particular, we are interested in robotic manipulation applications, where occlusions difficult accurate object pose estimation, and where behavior is dominated by contact interactions. 

\begin{figure}[!ht]
\centering
	\includegraphics[width=\linewidth]{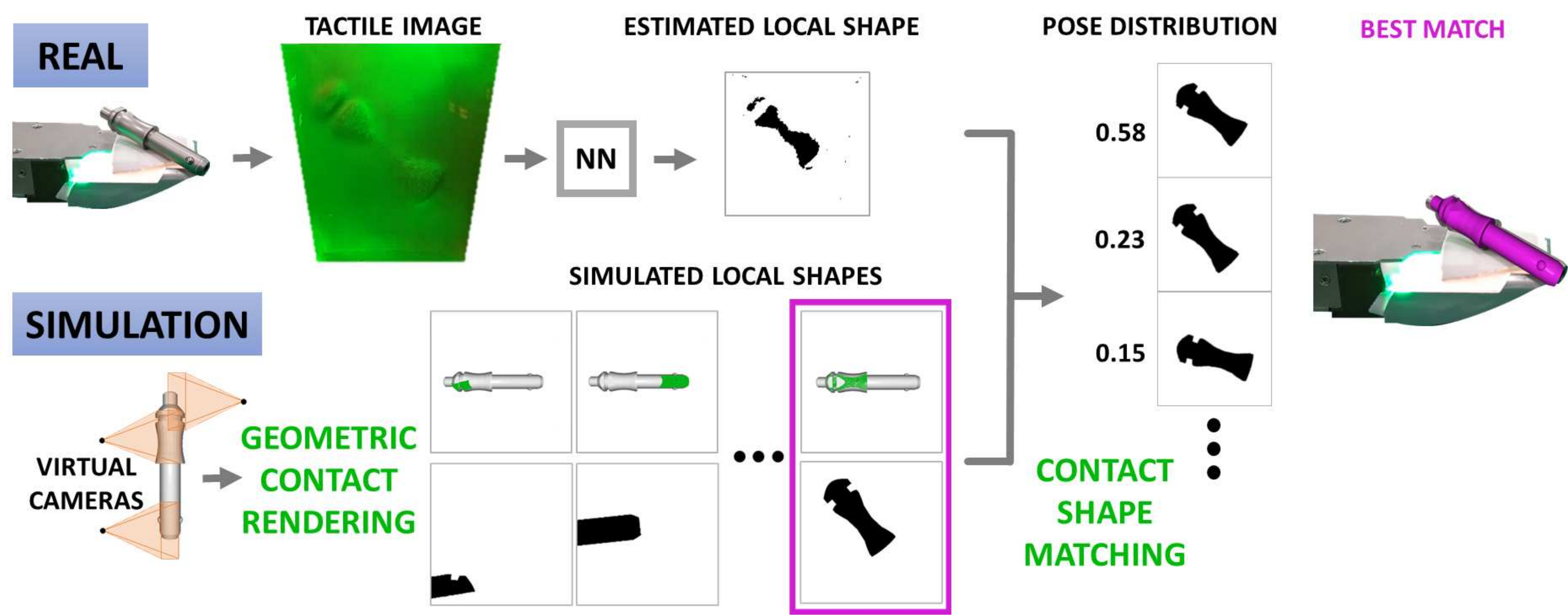}
\centering
\vspace{-7mm}
\caption{\textbf{Tactile pose estimation.} (Bottom row) In simulation, we render geometric contact shapes of the object from a dense set of possible contacts between object and tactile sensor. (Top row) The real sensor generates a tactile image from which we estimate its geometric contact shape. We then match it against the simulated set of contact shapes to find the distribution of contact poses that are more likely to have generated it. For efficiency and robustness, we do the contact shape matching in an embedding learned for that particular object. }\label{fig:motivation}
\vspace{-3mm}
\end{figure}

In this paper, we propose a framework to estimate the pose of a touched object, as illustrated in Figure~\ref{fig:motivation}.
Given a 3D model of the object, our method learns an object-specific perception model in simulation, tailored at estimating the pose of the object from one--or possibly multiple--tactile images. 
As a result, the approach localizes the object from the first touch, i.e., without requiring any previous interaction. 
The perception model is based on merging two key ideas:
\begin{itemize}
    \item \textbf{Geometric contact rendering:} we use the object model to render the contact shapes that the tactile sensor would observe for a dense set of contact poses. 
    
    \item \textbf{Contact shape matching:} given the estimated contact shape from a tactile observation, we match it against the precomputed dense set of simulated contact shapes. 
    The comparison happens in an object-specific embedding for contact shapes learned in simulation using contrastive learning tools~\citep{moco}. This provides robustness and speed compared to other methods based on direct pixel comparisons.\looseness=-1
\end{itemize}

Accounting by the discriminative power of tactile sensing, the proposed approach is motivated by scenarios where the main requirement is estimation accuracy and where object models will be available beforehand. 
Many industrial scenarios fit this category.

Most previous solutions to tactile pose estimation require prior exploration of the object~\citep{li_2014,bauza2019tactile}.
Acquiring this tactile experience can be expensive, and in many cases, unrealistic. 
In this paper, instead, we learn the perception model directly from the object geometry. 
The results in Sec.~\ref{sec:results}  show that the model learned in simulation directly transfers to the real world.
We attribute this both to the object-specific nature of the learned model and to the high-resolution nature of the tactile sensors used. \looseness=-1

Also, key to the approach is that, by simulating a dense set of tactile imprints, the algorithm can reason over pose distributions, not only the best estimate.
The learned embedding allows us to efficiently compute the likelihood of each contact shape in the simulated dense set to match with the predicted contact shape from the tactile sensor. This results in a probability distribution over object poses rather than just a single pose estimate.
Predicting distributions is key given that tactile sensing provides local observations, which sometimes might not be sufficiently discriminative. 

Finally, by maintaining probability distributions in pose space, we can incorporate extra constraints over the likelihood of each pose.
We illustrate it in the case of multi-contact, where information from multiple tactile observations must be combined simultaneously. 
By operating in a discretization of the pose space, the framework can potentially handle other pose constraints including those coming from other perception systems (e.g., vision), previous observations, or kinematics. 

In summary, the main contribution of this work is a framework for tactile pose estimation for objects with known geometry, with the following primary strengths:
\begin{itemize}
    \item[1.] Provides accurate pose estimation from the first touch, without requiring any previous interactions with the object. 
    \item[2.] Reasons over pose distributions by efficiently computing probabilities between a real contact shape and a dense set of simulated contact shapes.
    \item[3.] Integrates pose constraints, such as those arising from multi-contact scenarios where multiple observations and sensor poses must be considered.
\end{itemize}

\vspace{-2mm}
\section{Related Work}
\vspace{-2mm}
Tactile perception has been extensively explored in the robotics community. Relevant to this paper, this has resulted in the development of high-resolution tactile sensors and their use in a wide range of robotic manipulation applications. In this section, we review works that study tactile pose estimation and refer the reader to~\citep{Luo2017review} for a more in-depth review of tactile applications. 

While we propose to use high-resolution tactile sensors that are discriminative and rich in contact information, most initial works in tactile localization were meant for low-resolution tactile sensors~\citep{schaeffer2003methods,corcoran2010tracking,petrovskaya2011,chalon2013online,Bimbo2016,saund2017touch,javdani2013efficient, chebotar2014learning}. Moreover, in many cases, the sensors are bulky, or the objects considered are planar or consist of simple geometries. 
Finally, some works explore how to combine multiple tactile readings and reason in the space of contact manifolds~\citep{Koval2017, koval2015}. However, these are still based on low-resolution tactile feedback, often a binary contact/no-contact signal, and require many tactile readings to narrow pose estimates.

Given the challenges from the locality of tactile sensing, recent works have gravitated towards two different approaches. Combining tactile and vision to obtain better global estimates of the object pose or using higher-resolution tactile sensors that can better discriminate different contacts. Among the solutions that combine vision and tactile, most rely on tactile sensors as binary contact detectors whose main purpose is to refine the predictions from vision~\citep{bimbo2015global, Allen1999,Ilonen2014, Falco2017,yu2018realtime}. 

Other works, more in line with our approach, have focused on using high-resolution tactile sensors as the main sensing source for object localization. Initial works in this direction
used image-based tactile sensors to recover the contact shape of an object and then use it to filter the object pose~\citep{platt2011using,pezzementi2011, Luo2017}. However, these approaches only provide results on planar objects and require previous tactile exploration.

There has also been some recent work on highly deformable tactile sensors for object localization~\citep{Kuppuswamy2019FastMC}.  These sensors are large enough to fully cover the touched objects, which eases localization but limits any complex manipulation. 
In this work, we use the image-based tactile sensor GelSlim~\citep{donlon2018gelslim}. The sensing capabilities of these high-resolution sensors have already proven useful in multiple robotic applications, including assessing grasp quality~\citep{Hogan2018}, improving 3D shape perception~\citep{Wang2018} or directly learning from tactile images how to do contour following~\citep{lepora2019pixels} or tactile servoing~\citep{tian2019manipulation}.\looseness=-1

For the task of tactile object localization, \citet{li_2014} proposed to extract local contact shapes from objects to build a map of the object and then use it to localize new contacts. The approach is meant to deal with small parts with discriminative features. Later \citet{izatt_2017} proposed to compute pointclouds from the sensor and use them to complement a vision-based tracker. Their tracker is unimodal and cannot deal with the uncertainty that arises from the locality of tactile sensing. Finally, in previous work~\citep{bauza2019tactile} we proposed to extract local contact shapes from the sensors and match them to the tactile map of the objects to do object pose estimation.  This approach 
requires the estimation of a tactile map for each object by extensively exploring them with the sensor. 

In comparison, our approach moves all object-specific computations to simulation and only requires an object-agnostic calibration step of the tactile sensor to predict contact shapes. As a result, we can render in simulation contact shapes and learn object-specific models for pose estimation that translate well to the real world and achieve good accuracy from the first contact. 

Finally, our approach to tactile pose estimation is related to methods recently explored in the computer vision community where they render realistic images of objects and learn how to estimate the orientation of an object given a new image of it~\citep{sundermeyer2018implicit,Jeon2020}. 
While in vision, the most likely estimate is often sufficient, in tactile sensing, different object poses are more likely to produce the same observation. 
To address this problem, we explicitly reason over pose distributions when assessing the performance of our perception models. 

\vspace{-2mm}
\section{Method}
\vspace{-3mm}
%

We present an approach to object pose estimation based on tactile sensing and known object models; illustrated in Fig.~\ref{fig:motivation}.
In an object-specific embedding, we match a dense set of simulated contact shapes against the estimated contact shape from a real tactile observation. 
This results in a probability distribution over contact poses that can be later refined using other pose constraints or registration techniques.\looseness=-1

The algorithm starts from a geometric model of the object, and a description of the local geometry of a region of the object, a.k.a. \emph{contact shape}, captured by a tactile sensor.
In the case of this paper, we predict real contact shapes directly from the raw tactile images that the sensor outputs (\secref{sec:tactile_mapping}).

The next steps of our approach exploit the object model to estimate the object pose and are learned in simulation without using any real tactile observations. 
First, we develop \textit{geometric contact rendering}, an approach to simulate/render contact shapes in the form of images using the object model (\secref{sec:tactile_rendering}). 
Next, we generate a dense set of contact poses and their respective contact shapes, and use contrastive learning to match contact shapes depending on the closeness of their contact poses (\secref{sec:pose_estimation}). 
As a result, given the estimated contact shape from a real tactile observation, we can match it against this pre-computed dense set to obtain a probability distribution over contact poses. 

To predict contact poses beyond the resolution of the pre-computed set, we combine our approach with registration techniques on the contact shapes (\secref{sec:refinement}). 
Finally, we show our perception model is not restricted to single tactile observations and can handle multi-contact scenarios, and additional pose constraints (\secref{sec:multi_contact}).

\subsection{Contact shape prediction from tactile observations}\label{sec:tactile_mapping}

Given a tactile observation, our goal is to extract the contact shape that produces it. 
To that aim, we train a neural network (NN) that maps tactile observations to contact shapes following the approach we proposed in~\citet{bauza2019tactile}. 
The input to the NN is a normalized rescaled RGB tactile image of size 200x200. The output corresponds to a normalized one-channel depth image of size 200x200 that represents the contact shape. 

The training data is collected autonomously in a controlled 4-axis stage that generates controlled touches on known 3D-printed shapes.
Note that, for each tactile sensor, we only need to gather calibration data once because the map between tactile observations and contact shapes is object-independent. 
We provide further implementation details in ~\secref{sec:append_label} of the appendix.

\subsection{Contact shape rendering in simulation}\label{sec:tactile_rendering}

Given the geometric model of an object and its pose w.r.t. the sensor, we use rendering techniques to simulate local contact shapes from object poses. We refer to this process as \textit{geometric contact rendering}.
Below, we describe how we compute object poses w.r.t. the sensor that would result in contact without penetration, and their associated contact shapes:

\begin{enumerate}
    \item We create a rendering environment using the open-source library Pyrender~\citep{pyrender}. In this environment, we place a virtual depth camera at the origin looking in the positive z-axis. The sensor can be then imagined as a flat surface (a  rectangle in our case) orthogonal to the z-axis and at a positive distance $d$ from the camera. 
    \item We place the object in any configuration (6D pose) such that all points in the object are at least at a distance $d$ in the z-axis from the origin. 
    \item Next, we compute the smallest translation in the z-axis that would make the object contact the surface that represents the sensor. 

    \item Finally, we move the object accordingly and render a depth image. Its smallest pixel value corresponds to a depth $d$, and we consider that only pixels between distances $d$ and $d + \Delta d$ are in contact with the sensor. The rest are marked as non-contact. 
    The resulting depth image corresponds to the simulated contact shape of the object at that particular pose.
\end{enumerate}

As a result, given a contact pose, we can easily compute its corresponding simulated contact shape by rendering a depth image from the object mesh. 
For our sensor, depth images have a width and height of 470x470 (later rescaled to 200x200 for faster to compute), the distance to the origin is $d=25$mm, and the contact threshold is $\Delta d = 2$mm. 
The intrinsic parameters of the virtual camera are $f_x =291.5$, $f_y = 289$, $c_x = 235$ and $c_y = 235$. 

\begin{figure}[t]
\centering
	\includegraphics[width=\linewidth]{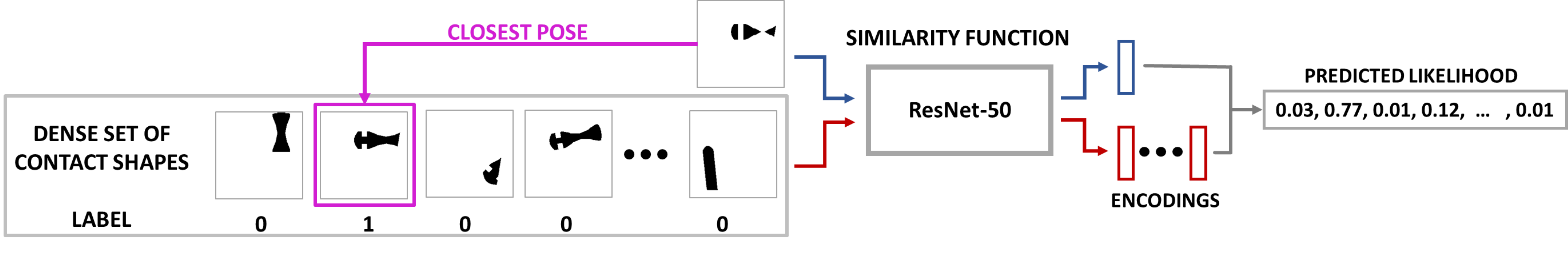}
\centering
\vspace{-7mm}
\caption{\textbf{Similarity function.} We build a similarity function that learns to encode contact shapes into a low dimensional space and predicts, given a new contact shape, the likelihood of being the closest match of each contact shape in the pre-computed set. By learning an encoder for the contact shapes, we can compare them very efficiently.
} \label{fig:encoder}
\vspace{-5mm}
\end{figure}

\subsection{Global tactile pose estimation}\label{sec:pose_estimation}

Once we know how to compute contact shapes both in simulation and from real tactile imprints, we reduce the problem of object pose estimation to finding what poses are more likely to produce a given contact shape. 
We solve this problem by first discretizing the space of possible contact poses as a parametrized grid, and then learning a similarity function that compares contact shapes.

\myparagraph{Object-dependent grids.} \label{sec:grid} Using the 3D model of an object, we discretize the space of object poses in a multidimensional grid. 
Building a grid in the space of poses is a well-studied problem~\citep{yershova2010generating,rocsca2014new} that makes finding nearby neighbors trivial. 
It also allows each point on the grid to be seen as the representative of a volumetric part of the space which helps to reason over distributions. 
We prune the grid by only keeping the poses that result in contact, and pair each of them with their respective contact shapes.
Since we only consider poses that result in contact, the z-dimension is fully determined by this condition, which results in 5-dimensional grids. 
Using a discrete structured set of poses allows us to easily account for object symmetries which can significantly reduce the grid size. \looseness=-1

\myparagraph{Similarity metric for contact shapes.} Given a new contact shape, we want to compare it to all pre-computed contact shapes in the grid to find what poses are more likely to produce it. 
To that aim, we modify MoCo~\citep{moco}, a state-of-the-art algorithm in contrastive learning, to encode contact shapes into a low dimensional embedding based on their pose distance. 
Compared to the original MoCo algorithm, the elements in the queue are fixed and assigned to each of the poses in the object's grid. Given a new contact shape, our model predicts the likelihood that each pose in the grid has produced the given shape. \looseness=-1 

To implement the encoder, we also use a ResNet-50~\citep{he2016}, but cropped before the average-pooling layer to preserve spatial information, making it a fully-convolutional architecture. 
The loss function is the Cross-Entropy loss which allows us to predict probabilities. 
The training data comes from selecting a random contact pose and finding its closest element in the dense grid. 
Then, we use as desired probabilities a vector of all zeros except for the closest element which gets assigned to probability one (see Fig.~\ref{fig:encoder}).
%
%
%
~\secref{sec:append_sim} in the appendix contains further details on the learning method.\looseness=-1


Once we have created a dense grid and trained a similarity encoder for an object, given a new contact shape, we can estimate which poses from the grid are more likely to generate it.
To run our method in real-time, we first encode the given contact shape (0.0075s, 133Hz) and then compare it to all pre-computed encodings from the grid, which requires a single matrix-vector multiplication. Finally, we perform a softmax over the resulting vector to obtain a probability distribution over contact poses

\subsection{Contact shape refinement}\label{sec:refinement}

To avoid limiting pose estimations by the resolution of the grid, we refine the most likely poses and their contact shapes to better match the contact shape we are interested in. 
Because contact shapes come from depth images, we can convert them to pointclouds and use registration techniques to find the best transformation between them. 
%

We evaluated several pointcloud registration algorithms in simulation and concluded that FilterReg~\citep{gao2019filterreg}, a state of the art registration algorithm twice as fast as ICP, performed the best even with just one iteration. 
Applying FilterReg between two contact shapes takes less than 0.01s (100Hz). 


\subsection{Multi-contact pose estimation}\label{sec:multi_contact}

In this section, we show how to extend our approach to multi-contact settings where we simultaneously need to reason over several tactile observations.  
In~\secref{sec:append_multi} in the appendix we show that the likelihood of an object pose from the grid, $x$, given the estimated contact shapes, $CS_{1,..., N}$, from N sensors is proportional to 
\begin{equation}
P(x | CS_1, ... , CS_N) \propto P(x | CS_1) \cdot ... \cdot P(x | CS_N) \label{eq:multi-contact}
\end{equation}
when there are no priors on object pose, i.e., all elements in the grid are equally likely, and the embedding network has been trained using uniformly-sampled poses. $P(x | CS_i)$ is the likelihood that pose $x$ produces the contact shape $CS_i$ on sensor $i$. 
When a prior over poses is available, we can add a prior term, $P_{task}(x)$, to compute $P(x | CS_1, ... , CS_N)$ (see appendix~\secref{eq:multi-contact}). This allows combining our approach with additional pose constraints such as the ones coming from kinematics, previous tactile observations, or other perceptions systems. 
%

The terms $P(x | CS_i)$ come directly from computing the similarity function between $CS_i$ and the contact shape from the grid of sensor $i$ that is closest to the contact pose $x$.
As a result, the computational cost of considering multiple contacts scales linearly with the number of sensors.\looseness=-1

\section{Results}
\vspace{-3mm}
\label{sec:results}
\subsection{Real data collection}

While most computations of the algorithm are done in simulation, the end goal of our approach is to provide accurate pose estimation in the real world. To that aim, we designed a system that collects  tactile observations on accurately-controlled poses. Below we describe the tactile sensor, the robot platform, and the objects used to perform the experiments.

\myparagraph{Tactile sensor.} We consider the tactile sensor GelSlim~\citep{donlon2018gelslim} which provides high-resolution tactile readings. The sensor consists of a membrane that deforms when contacted and a camera that records the deformation. The sensor publishes tactile observations through ROS as  470x470 compressed images at a frequency of 90Hz. 
Some regions of the image are masked because they do not record the deformed membrane. 

\myparagraph{Robot platform.} 
To get calibrated real pairs of contact poses and tactile observations, we fix the sensor to the environment and use a 4-axis robotic stage with translation and rotation in the horizontal plane and vertical motion. 
The platform is composed of 3 Newmark linear stages, two horizontal (x, y-axis), and one vertical (z-axis), and a rotational stage in the vertical direction. 
We use ROS to control the setup, which moves with submillimeter precision and high repeatability. 
To calibrate the sensor w.r.t. the platform, we use a set of marks at known poses to make sure the tactile images from touching them match the simulated contact shapes at these poses. 
We provide further details on the collection of labeled real data in~\secref{sec:append_collect} of the appendix.



\begin{figure*}[t]
\centering
	\includegraphics[width=0.9\linewidth]{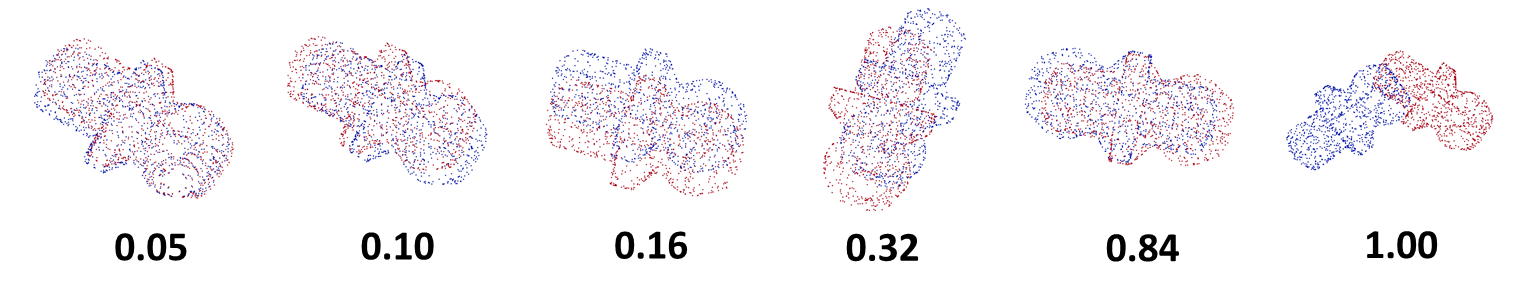}
\centering
\vspace{-5mm}
\caption{\textbf{Normalized pose errors}, i.e., pose errors w.r.t. to the average random error, for the \textit{damping pin}. The first case corresponds to the average closest distance in its grid, the 3rd to the median results from \textit{Best-1}, and the last one to the average random error. Finally, the example with 0.84 normalized error depicts a non-unique contact shape, i.e., the two object poses result in very similar contact shapes that are not possible to distinguish without additional information.\looseness=-1
} \label{fig:pose_error} 
\vspace{-5mm}
\end{figure*}

\myparagraph{Objects.}
We test our algorithm on 4 objects from the McMaster dataset~\citep{corona2018pose} (Figure~\ref{fig:pose_error}). We scaled up \textit{pin} and \textit{damping pin} to make them bigger and thus more challenging. 
%
%
%
In the video linked to this paper, we also included the objects \textit{cotter} and \textit{hook}, which are also from McMaster, to show additional qualitative results on the generalization of the algorithm.

For each object, we build a dense grid that contains the set of poses that would result in contact with the sensor. The distance between  closest neighbors is no larger than 2mm on average.  For the quantitative results, we restrict these grids to one face of the object which results in grids with 5k to 20k elements depending on the object dimensions.


\begin{figure}[t]
\centering
	\includegraphics[width=\linewidth]{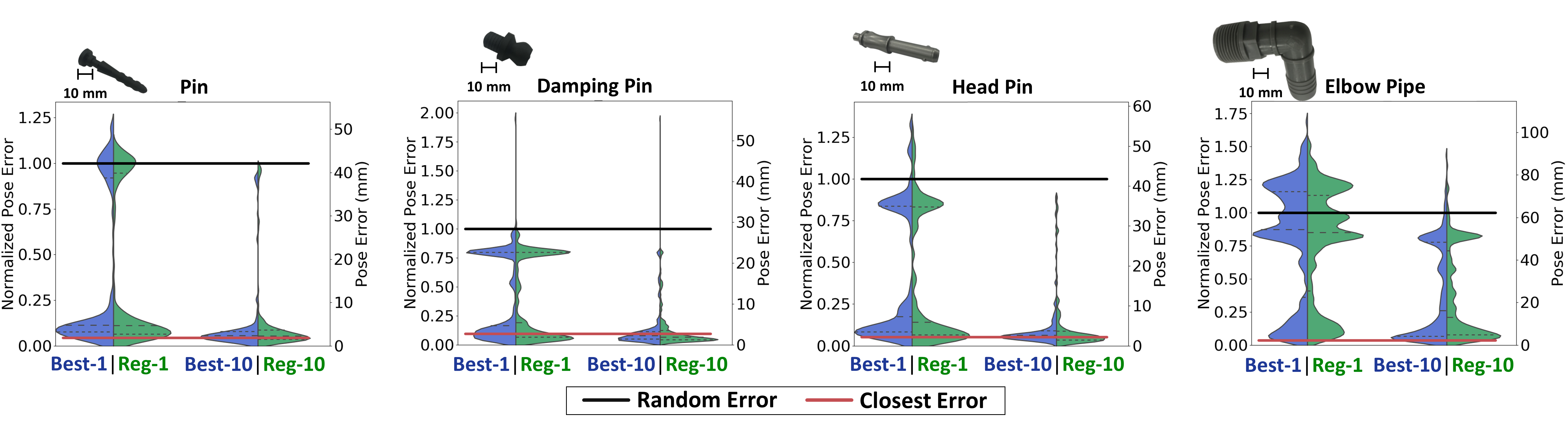}
\centering
\vspace{-7mm}
\caption{\textbf{Pose estimation results.} We show in blue the error distributions for the best match and the best out of 10. Green distributions show the pose error after also applying the pointcloud registration. Most of the distributions are far below random error (black line) and close to the error obtained when selecting the closest element from the grid (red line). For some objects like \textit{damping pin}, we see multimodality in the error distributions due to different contact poses resulting in similar contact shapes.\looseness=-1} \label{fig:pose_results} \label{fig:objects}
\vspace{-5mm}
\end{figure}

\subsection{ Real pose estimation results}

First, we test the accuracy of our approach at estimating object poses from single tactile imprints. 
For each object, we collected at least 150 pairs of tactile images and object poses. 
Given two poses, we measure their distance by sampling 10K points on the object 3D model and averaging the distance between these points when the object is at either of the two poses. 
This distance is sometimes called ADD (average 3D distance) but, for simplicity, we just refer to it as the pose error.

To be able to compare errors across shapes and object sizes, we also compute the \textit{normalized pose error} which divides the original pose error by the average error obtained from predicting a random contact pose. Figure~\ref{fig:pose_error} shows examples of different normalized pose errors.


Figure~\ref{fig:pose_results} shows the accuracy results for tactile pose estimation for each of the four objects, in the form of error distributions. The black line in each plot represents the random error, i.e., the average error when predicting a random contact pose. The red line measures the average pose error between the ground truth pose and its closest pose in the grid. This sets a lower bound on the average performance when only considering poses from the grid, without point-cloud registration. 

For each object, we include the error distributions for:
\begin{enumerate}
    \item \textit{Best-1}: only considers the most likely pose of the grid.
    \item \textit{Reg-1}: refines the most likely pose using FilterReg.
    \item \textit{Best-10}: considers the 10 most likely poses of the grid and selects the one that leads to the lowest pose error. This approach requires knowledge of the true pose and it is not applicable in practice. However, it helps to understand the quality of the predicted pose distributions, which is especially relevant when different object poses can lead to very similar contact shapes.\looseness=-1
    \item \textit{Reg-10}: takes the 10 most likely poses of the grid, refines them using FilterReg, and selects the one with the lowest error. Again, this is not a viable approach but a measure of the quality of the pose distributions. 
\end{enumerate}



For all objects, we observe that most of the error distributions are below the expected random error. 
Moreover, a large fraction of the errors are small and gathered around the closest error from the grid (red line). Fig.~\ref{fig:pose_rand_results} from the appendix compares these error distributions against the one obtained from randomly selecting contact poses.
In some cases, the error distributions are multimodal. 
This happens especially when there is non-uniqueness, i.e, different poses of the object lead to very similar contact shapes. 
These poses exist for all 4 objects, but are most important in the \textit{damping pin} and the \textit{elbow pipe}. 
Figure~\ref{fig:pose_error} shows an example for the \textit{damping pin} of non-uniqueness in the contact shape in the example with a normalized error of 0.84. This error matches the second mode in the error distributions for this object.

Selecting the best error out of the 10 best poses results in considerably lower errors and suggests that our approach can provide useful pose distributions. Multimodality also disappears or gets reduced in these distributions, meaning that we can capture non-unique cases. 
The case of the \textit{elbow pipe} is especially interesting because it has multiple non-unique poses. Therefore, from a single tactile observation, it is often not possible to predict its pose. Because our method provides distributions, we can capture the non-unique cases when considering several pose predictions instead of a single estimation. If we observe the best out of 10, the error distribution becomes much closer to zero, meaning that with 10 samples we can already capture several poses that result in very similar contact shapes.\looseness=-1  

Another important observation coming from these plots is that the median errors are low even when only considering the most likely pose. 
For \textit{Best-1}, we get median normalized errors of 0.11 for \textit{pin} (4.8mm), 0.16 for \textit{damping pin} (4.6mm), and 0.18 for \textit{head} (7.3mm). 
The \textit{elbow pipe} is more challenging, and we get median normalized errors of 0.87 (54.4mm), however, if we consider the \textit{Best10} we get 0.26 (16.3mm), and 0.08 (5mm) for \textit{Best50} which represents selecting only 50 poses out of more than the 20k in its grids. 
Finally, adding FilterReg further improves the results. This is because it can refine the pose estimates when the initial distance between contact shapes is small enough and locally transforming them is possible. 

\begin{figure}[t]
\centering
	\includegraphics[width=\linewidth]{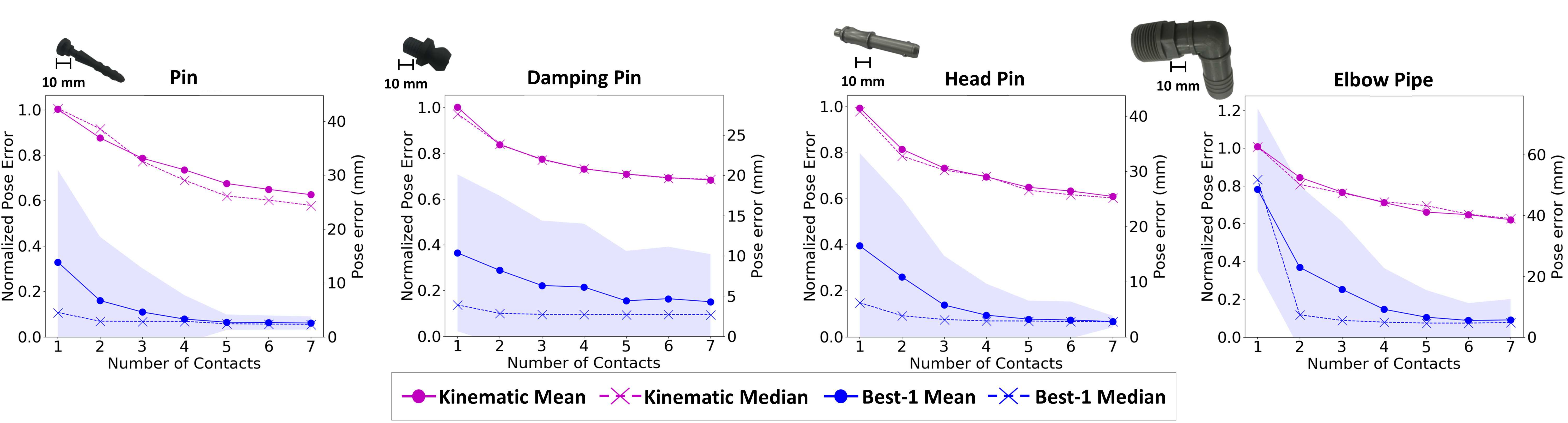}
\centering
\vspace{-6mm}
\caption{\textbf{Multi-contact results.} We compare the median and mean errors for \textit{Best-1} and \textit{kinematic}. \textit{Kinematic} does not deliver good results even after many contacts, suggesting that using contact shape information is key for \textit{Best-1} to achieve good results. 
The predictions from \textit{Best-1} improve  with the number of contacts. The mean and median become closer while the standard deviation also decreases, confirming that our approach can successfully integrate multi-contact information. }
\vspace{-5mm}
\label{fig:multi_results}

\end{figure} 

\subsection{ Multi-contact pose estimation results}

Next, we evaluate our solution to pose estimation using multiple contacts. 
For each object, we evaluate our approach on 100 examples. 
We synthesize multi-contact examples by combining data from single contacts. 
For each example we add, one by one, pairs of tactile images and sensor-object poses to increase the number of contacts from 1 to 7.  

We assess the performance of our pose estimations by reporting both \textit{Best-1}, the best match from the grid, and \textit{kinematic}, which only considers the kinematic constraints imposed by the sensors poses and whether the sensor is in contact or not. 
Figure~\ref{fig:multi_results} compares both approaches using the normalized error and the original pose error. 
We observe that only using kinematic information is not enough to substantially reduce the uncertainty in object pose. 
This is because only detecting if there is contact does not inform on what part of the sensor is being touched which results in many poses remaining possible even after multiple contacts.

In contrast, for our approach \textit{Best-1}, adding more contacts results in lower mean and median errors, and the standard deviation decreases. In the case of \textit{elbow pipe}, just adding one additional contact already reduces dramatically the pose error.
This reinforces the idea that adding sensed contacts helps to reduce the ambiguities coming from having different poses that result in similar contact shapes.\looseness=-1

\vspace{-2mm}
\section{Discussion}
\vspace{-2mm}
This paper presents an approach to tactile pose estimation for objects with known geometry. For tactile sensors that provide local contact shape information, we learn in simulation how to match real estimated contact shapes to a precomputed dense set of poses by rendering their contact shapes. The approach allows to reason over pose distributions and to handle additional pose constraints.

Our approach relies on learning an embedding that facilitates comparing simulated and real contact shapes. However, real contact shapes come from a distribution that differs from the training data. To bridge this gap, we rely on improving the methods that extract contact shapes and explore data augmentation techniques to learn embeddings that better accommodate the discrepancies between real and simulated contact shapes. 

Similarly, we assume we have access to accurate geometric models of objects. This makes our approach applicable to many real scenarios like assembly automation but also limits its use-cases. Our solution could effectively be combined with reconstructed object models, which would require addressing the combined challenges of pose and shape uncertainty. This might demand to learn contact shape embeddings that adapt to both. We believe the idea of matching estimated contact shapes to a dense precomputed set opens the door to moving many computations into simulation and improving how robots learn to perceive and manipulate their environment.



\clearpage
\acknowledgments{We thank Ferran Alet for providing detailed insights and carefully reviewing the paper. We also thank Siyuan Song and Ian Taylor for helping with the real setup, and Yen-Chen Lin for helpful discussions.  This work was supported by the Toyota Research Institute (TRI). This article solely reflects the opinions and conclusions of its authors and not TRI or any other Toyota entity. Maria Bauza is the recipient of a Facebook Fellowship.}   


\bibliography{biblio}  
\newpage
\appendix

\section{Contact shape prediction from tactile observations}\label{sec:append_label}

Given a tactile observation, our goal is to extract the contact shape that produces it. 
To that aim, we train a neural network (NN) that maps tactile observations to contact shapes following the approach proposed in~\citep{bauza2019tactile} and described below for completeness.

\begin{figure}[h!t]

\centering
	\includegraphics[width=\linewidth]{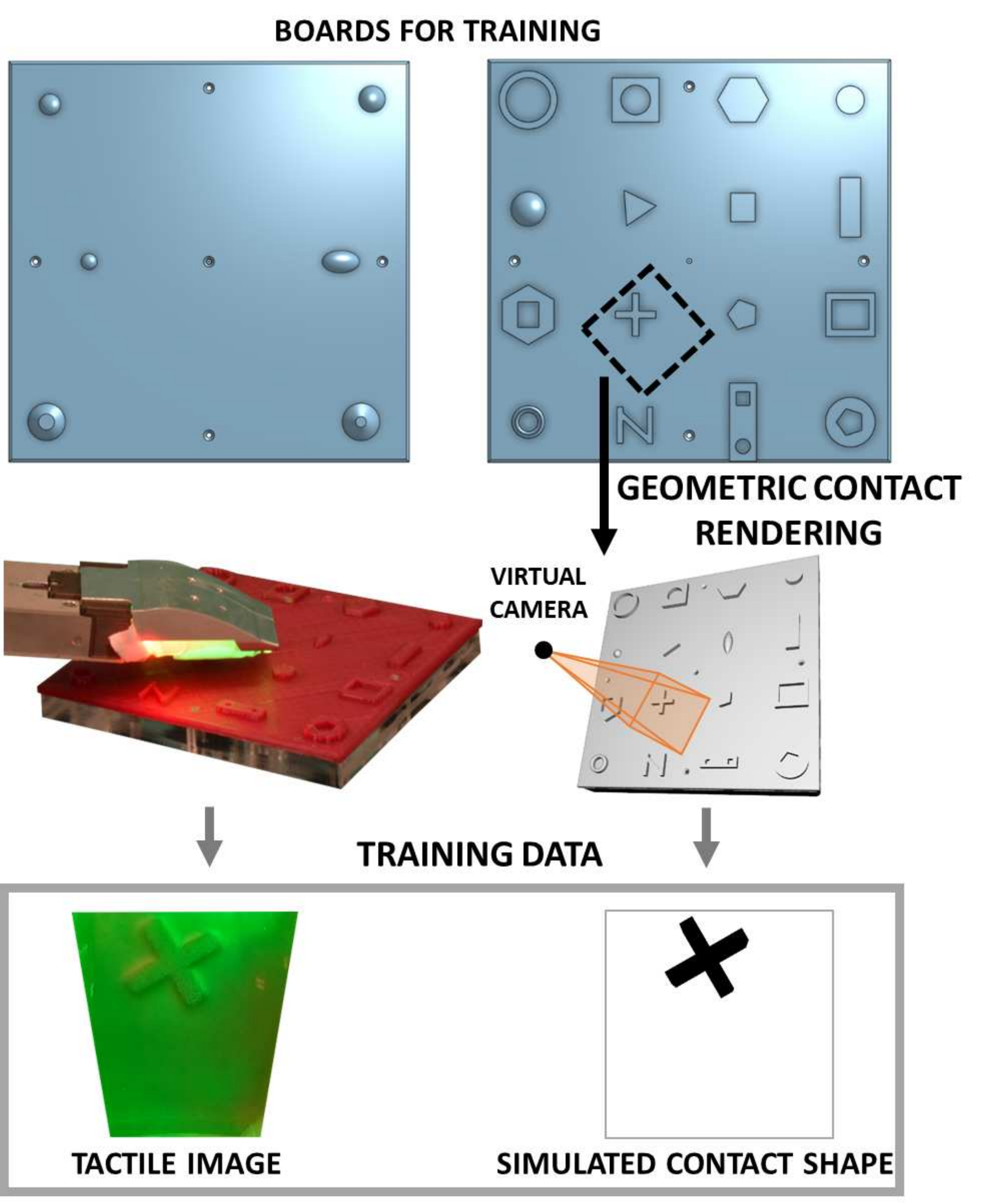}
\centering

\caption{\textbf{Training data set-up to estimate real contact shapes from real tactile observations.} We use the boards at the top to collect real tactile observations from them as well as simulated contact shapes. By placing the boards on a robotic platform, we can move them w.r.t. the sensor to perform calibrated touches on them. From each touch, we recover both a tactile image and a simulated contact shape that we compute using the 3D model of the boards and its pose w.r.t the sensor when the touch happened.} \label{fig:data_collection}
\vspace{-5mm}
\end{figure}

As shown in Fig.~\ref{fig:data_collection}, the input to the NN is a normalized rescaled RGB tactile image of size 200x200. The output corresponds to a normalized one-channel depth image of size 200x200 that represents the contact shape.\looseness=-1

The training data is collected autonomously in a controlled 4-axis stage that generates controlled touches on known 3D-printed shapes. In our case, we collect the calibration data from two 3D-printed boards with simple geometric shapes on them (see the top of Fig.~\ref{fig:data_collection}). From each controlled touch, we obtain a tactile observation and the pose of the board w.r.t. to the sensor. From this pose, we can later simulate the corresponding contact shape to the tactile observation using geometric contact rendering and the 3D model of the board.

Note that, for each tactile sensor, to do tactile localization on any object, we only need to gather calibration data once because the map between tactile observations and contact shapes is object-independent. 

Given the training data collected from our sensor, we normalize tactile images (input) and contact shapes (output) using the mean and standard deviation of the training set. 
Before normalization, the contact shape values range between 0 and 2 mm, where zero means maximum penetration into the sensor, and 2 mm indicates no contact. 

The architecture of the NN consists of 6 convolutional layers with ReLU as the activation function and a kernel size of 3 except for the last layer which is 1. Between layers 5 and 6, we add a Dropout layer with a fraction rate of 0.5. The optimizer is Adam with a learning rate of 0.0001, and the loss function is the RMSE between pixels. 

In practice, we collected 1000 pairs of tactile observations and contact shapes for each board, and trained the NN for 30 epochs. Each forward pass on the NN takes less than 0.005s (200Hz).

\subsection{Collecting ground-truth data for real objects}\label{sec:append_collect}

\begin{figure}[ht]

\centering
	\includegraphics[width=\linewidth]{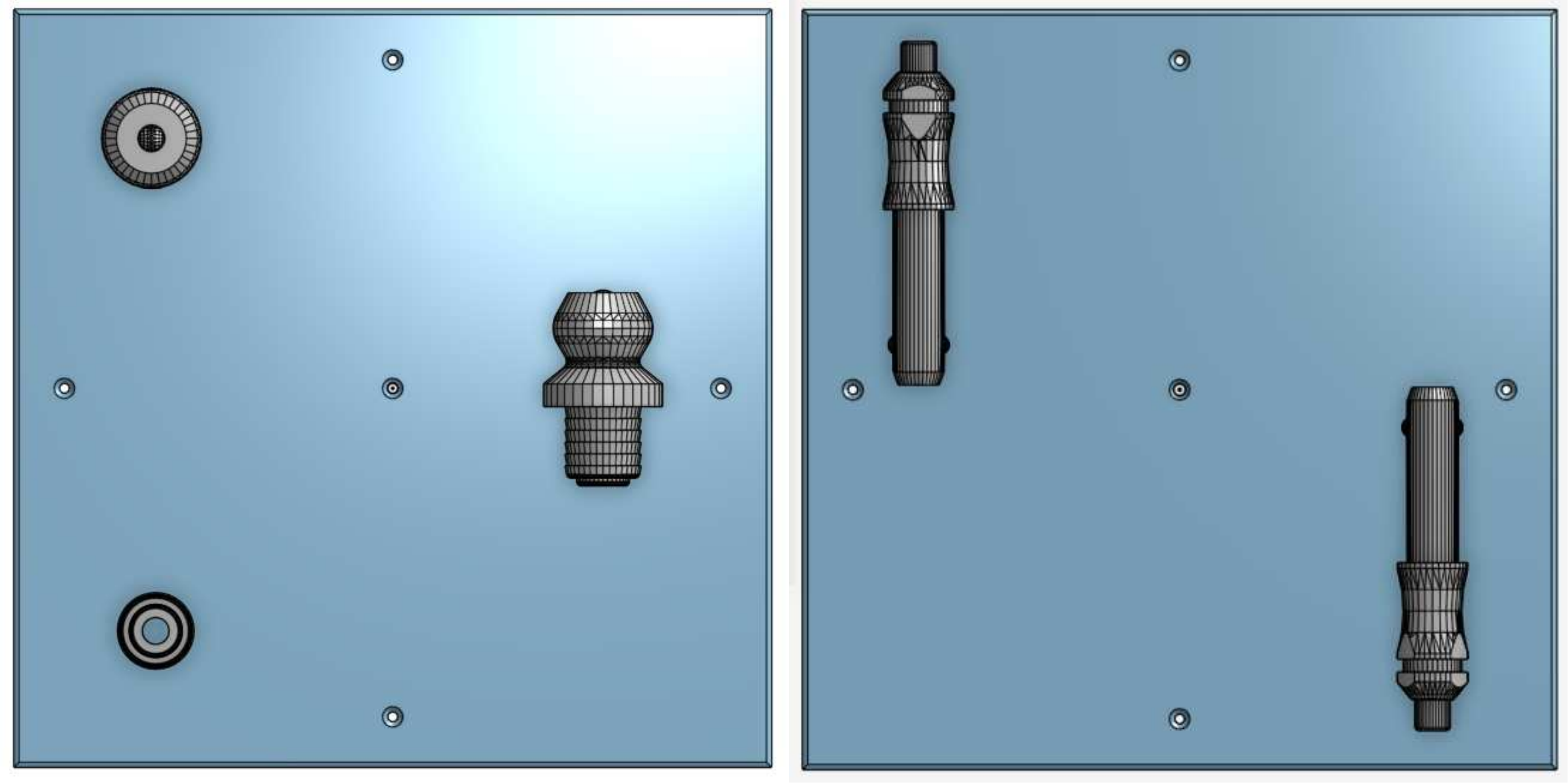}
\centering

\caption{\textbf{Boards used for ground-truth data collection.} We sliced the models of the objects and create boards with them. As a result, we can 3D-print these boards and use them to collect ground-truth data for test objects.} \label{fig:test_data}
\vspace{-5mm}
\end{figure}

To evaluate our approach to tactile localization, we collected ground-truth data from novel objects using the same set-up. 
The resulting data from the set-up consists of tactile observations paired with the calibrated pose of the object.
This allows us to compute pose errors between our method predictions from just the tactile observations, and the true calibrated object poses.
Figure~\ref{fig:test_data} shows two of the boards used to collect ground-truth data on the real objects. 

The results displayed in the paper's video show examples of tactile images and the predicted contact shapes obtained using the NN that goes from tactile observations to contact shapes. We note that this NN has only been trained once using only calibration data from the calibration boards in Fig.~\ref{fig:data_collection}.

\section{Similarity metric for contact shapes}\label{sec:append_sim}

Given a new contact shape, we want to compare it to all pre-computed contact shapes in the grid to find what poses are more likely to produce it. 
To that aim, we use MOCO~\citep{moco}, a state-of-the-art tool in contrastive learning, to train a NN that encodes contact shapes into a low dimensional embedding allowing to compare them using the distance between their encodings.

The NN encoder is based on a pre-trained Resnet-50~\citep{he2016} cropped before the average-pooling layer to preserve spatial information. It encodes contact shapes into vectors of dimension 3200.
To train, we use the same optimizer as~\citet{moco}: stochastic gradient descent with a learning rate that starts at 0.03 and decays over time, momentum of 0.07, and weight decay of 0.0001.
Each training datapoint comes from selecting a random contact pose, rendering its contact shape, and finding its closest pose in the pre-computed grid. Then our input is the contact shape. The output or label is a vector with all entries zeros except the one that corresponds to the closest element that gets a one. This vector represents the probability of each poses in the grid to be the closest to the target pose. 
The loss function is the cross-entropy loss between the label and the softmax between the target encoder and the ones from the grid (see Fig.~\ref{fig:encoder}). As the NN gets trained, we recompute the encoding of the queue following~\citep{moco}.
We train the models for 30 epochs, as performance does not significantly change after that. 

To make the training data resemble more the real distribution of contact shapes, we train using as target, contact shapes with different depth thresholds, $\Delta d$, selected randomly between 1 and 2 mm. This accounts for changes in the contact force applied to the sensor that can make contacts more or less deep (see Fig.~\ref{fig:encoder}).
Finally, before encoding the contact shapes, we converted them into binary masks to facilitate the learning process and prevent that numerical mismatches between real and simulated contact shapes affect the results.

\section{Multi-contact for N sensors}\label{sec:append_multi}

To derive Equation 1 in Section 3.5, we will prove a more general equation for the object pose $x$ when it is in contact with N sensors:

\begin{equation}
P(x | CS_1, ..., CS_N) = \frac{P(x | CS_1) \cdot ... \cdot P(x | CS_N) \cdot P_{task}(x)/ P_{train}(x)^{N}}{\sum_{x' }P(x' | CS_1) \cdot ... \cdot P(x' | CS_N) \cdot P_{task}(x')/ P_{train}(x')^{N}} \label{eq:full}
\end{equation}
where $P(x | CS_i)$ is the likelihood that pose $x$ has produced a contact shape $CS_i$ on sensor $i$. $P_{task}(x)$ is a prior over all possible contact poses that would result in contact with the N sensors in the current task. $P_{train}(x)$ is the distribution over the contact poses used to train the similarity function in ~\secref{sec:tactile_mapping}.

Next, we prove Equation~\eqref{eq:full}.

Assuming that we have access to the poses of the N sensors, given an object pose $x$, the contact shapes $\{CS_i\}$ become determined. This allows us to write the joint distribution of pose and contact shapes as:

$$
P(x, CS_1, ..., CS_N) = P_{task}(x) \cdot P(CS_1 | x) \cdot ... \cdot P(CS_N | x)
$$

Now we can use Bayes theorem on the $P (CS_i | x)$ terms:

$$
P(CS_i| x )  = P(x | CS_i) \cdot P( CS_i)/P_{train}(x)
$$

and obtain:

$$
P(x, CS_1, ..., CS_N) =  P(x| CS_1) \cdot P(CS_1) \cdot ... \cdot P(x| CS_N) \cdot P(CS_N) \cdot P_{task}(x)/ P_{train}(x)^N
$$

where we take into account that the $P(x|CS_i)$ have been trained using a predefined distribution of poses $P_{train}(x)$ that does not necessarily match $P_{task}(x)$.
Now, we can compute $P(x| CS_1, ... , CS_N)$ using again the Bayes theorem:

$$
P(x| CS_1, ... , CS_N) = \frac{P(x, CS_1, ... , CS_N)}{P(CS_1, ... , CS_N)} = \frac{P(x, CS_1, ... , CS_N)}{\sum_{x'}P(x',CS_1, ... , CS_N)}
$$
and thus
$$
P(x| CS_1, ... , CS_N) = \frac{\left(\Pi_i^N  P(x| CS_i) \cdot P(CS_i) \right) \cdot  P_{task}(x)/ P_{train}(x)^{N} }{\sum_{x'} \left( \Pi_i^N P(x'| CS_i) \cdot P(CS_i) \right) \cdot  P_{task}(x')/ P_{train}(x')^{N}}
$$
We can now cancel the terms $P(CS_i)$ that appear both in the numerator and denominator, getting:
$$
P(x| CS_1, ... , CS_N) = \frac{\left(\Pi_i^N  P(x| CS_i) \right) \cdot  P_{task}(x)/ P_{train}(x)^{N} }{\sum_{x'} \left( \Pi_i^N P(x'| CS_i) \right) \cdot  P_{task}(x')/ P_{train}(x')^{N}}
$$

This concludes the proof of Equation~\eqref{eq:full}. In practice, we discretize the set of contact poses $\{ x \}$ that result in contact with the N sensors using the grid over of sensor 1 (this is an arbitrary choice). Then, we use the transformation between sensor 1 and sensor $i$ to compute each $P(x| CS_i)$ using the closest pose to $x$ in the grid of sensor $i$. Note that often a pose that results in contact with sensor $1$ will not contact sensor $i$. In that case, we do not consider that pose as we are only interested in poses that contact the N sensors. Finally, because the grids are dense and structured, finding the closest pose to $x$ in a grid has a minor effect on performance and is fast to compute.

We can use Equation~\eqref{eq:full} to derive Equation~\eqref{eq:multi-contact} in~\secref{sec:multi_contact} under two extra assumptions. First, we observe that the denominator is constant because all contact shapes $CS_i$ are given. Next, if we have no prior over the contact poses both during the task and training, then $P_{task}(x)$ and $P_{train}(x)$ are constant, and that leads to Equation~\ref{eq:multi-contact} repeated here for completeness: 
$$
P(x | CS_1, ... , CS_N) \propto P(x | CS_1) \cdot ... \cdot P(x | CS_N)
$$.

\section{Comparing our method against random}\label{sec:pose_rand}

\begin{figure}[h!]
\centering
	\includegraphics[width=\linewidth]{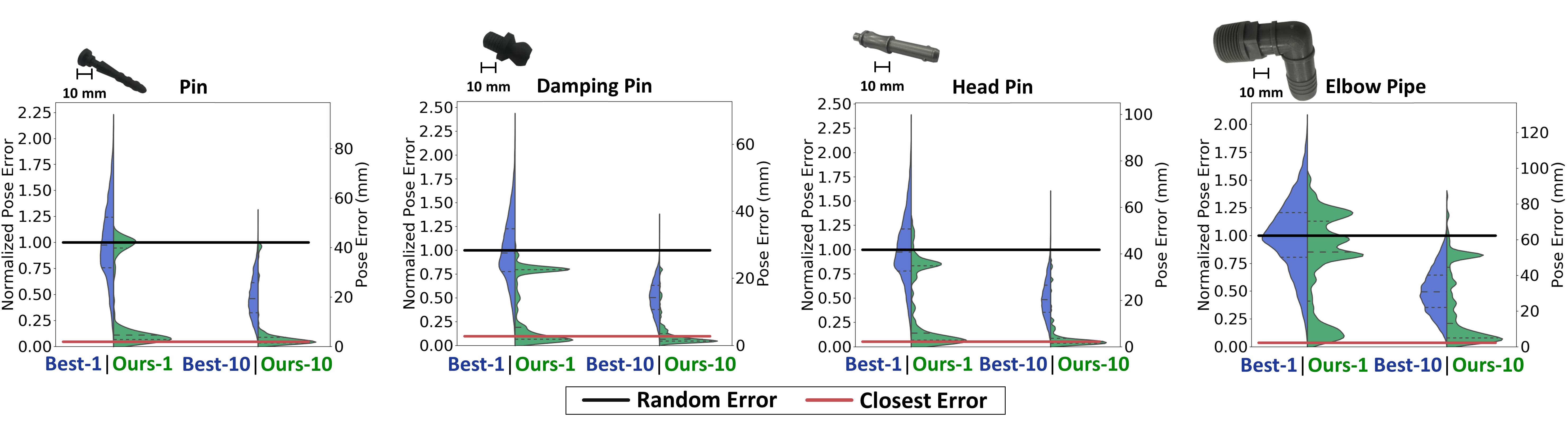}
\centering
\caption{\textbf{Pose estimation results vs. random.} We show in blue the error distributions when randomly selecting 1 or the best out of 10 random contact poses. Green distributions show the pose error of our method after applying the pointcloud registration. While the random distribution is wide and centered around its mean, our method provides much smaller errors and thus better localization. \looseness=-1}
\label{fig:pose_rand_results}
\end{figure}

\end{document}